\documentclass{article}

\usepackage{PRIMEarxiv}

\usepackage[utf8]{inputenc} 
\usepackage[T1]{fontenc}    
\usepackage{hyperref}       
\usepackage{url}            
\usepackage{booktabs}       
\usepackage{amsfonts}       
\usepackage{nicefrac}       
\usepackage{microtype}      
\usepackage{lipsum}
\usepackage{fancyhdr}       
\usepackage{graphicx}       
\graphicspath{{media/}}     

\title{A Compact and Semantic Latent Space for Disentangled and Controllable Image Editing
\thanks{\textit{\underline{DOI}}: 
\textbf{Gwilherm Lesné, Yann Gousseau, Saïd Ladjal, and Alasdair Newson. 2023. A Compact and Semantic Latent Space for Disentangled and Controllable Image Editing. In Proceedings of the 20th ACM SIGGRAPH European Conference on Visual Media Production (CVMP '23). Association for Computing Machinery, New York, NY, USA, Article 3, 1–10. https://doi.org/10.1145/3626495.3626508}} 
}

\author{
  Gwilherm Lesné, Yann Gousseau, Saïd Ladjal, Alasdair Newson\\
  LTCI\\
  Télécom Paris\\
  Palaiseau, FR\\
  \texttt{\{gwilherm.lesne, yann.gousseau, said.ladjal, anewson\}@telecom-paris.fr} \\
}

\def\etal{\emph{et al}}

\begin{document}

\maketitle

\begin{abstract}
\textit{
Recent advances in the field of generative models and in particular generative adversarial networks (GANs) have lead to substantial progress for controlled image editing, especially compared with the pre-deep learning era.
Despite their powerful ability to apply realistic modifications to an image, these methods often lack properties like disentanglement (the capacity to edit attributes independently). In this paper, we propose an auto-encoder which re-organizes the latent space of StyleGAN, so  that each attribute which we wish to edit corresponds to an axis of the new latent space, and furthermore that the latent axes are decorrelated, encouraging disentanglement. We work in a compressed version of the latent space, using Principal Component Analysis, meaning that the parameter complexity of our autoencoder is reduced, leading to short training times ($\sim$ 45 mins). Qualitative and quantitative results demonstrate the editing capabilities of our approach, with greater disentanglement than competing methods, while maintaining fidelity to the original image with respect to identity. Our autoencoder architecture simple and straightforward, facilitating implementation.}

\end{abstract}

\section{Introduction}

Since the advent of deep generative models, it has been possible to create random examples of synthetic, photorealistic images. Some of the most popular methods are Variational Auto-encoders~\cite{kingma2013auto}, normalizing flows~\cite{papamakarios2021normalizing}, diffusion models~\cite{song2020score} or Generative Adversarial Networks~\cite{goodfellow2020generative} (GANs). In particular, style-type GANs such as BigGAN \cite{BigGAN} or StyleGAN \cite{StyleGAN} have distinguished themselves for their capacity of high resolution synthesis. These generative models all rely on an internal \emph{latent space} which is learned to represent high-dimensional image data.
Given the synthesis power of these different methods, it is quite natural to use them for \emph{editing}. Indeed, once the latent space is learned, as has been observed in the literature, it is natural to suppose that moving a point in a well-chosen direction in the latent space will correspond to completing a high-level editing task. In the case of face images, for example, this will correspond to modifying attributes such as expression, age, glasses etc. of the face. The goal of deep generative model-based editing is to find this direction, given an input image and an editing goal. In order for such a method to be used in a practical setting by digital artists, several requirements must be met:
\begin{itemize}
    \item Disentanglement: by modifying one attribute, we do not modify any other attribute;
    \item Controllability: we require a direct, meaningful control over the attributes. In particular, we want to create a latent representation of an image where each axis corresponds to an attribute;
    \item Fidelity: when we edit an image, we do not wish to change the identity of a face or other non-labelled attributes, or navigate away from the original latent space.
\end{itemize}
Imposing all of these constraints simultaneously is the main challenge of image editing methods using deep generative networks. 

In this paper, we propose a method for image editing which targets these three goals. Our method is based on a re-organisation of the latent space of a GAN, by using an autoencoder. In this work, we concentrate on the StyleGAN2~\cite{StyleGAN2} model due to its great performance, however, the approach is very general and can be applied to any GAN. We start by observing that most latent spaces of GANs are over-parametrized, and can thus be further compressed. We do this using Principal Component Analysis \cite{GANSpace} carried out on StyleGAN2's latent space. We navigate only in this compressed latent space, which also makes sense if we wish to stay as close to the original latent space as possible, thus encouraging fidelity (as defined above). We then train an autoencoder to transform this compressed space to another latent space where each of the attributes corresponds to a given axis. This is achieved via a training loss function. We split up our new latent code in two parts: the first containing the attributes which we want to edit, and the second containing free components that allow for diversity in the image (face identity, lighting etc). We impose disentanglement between the axes of the first code by minimising their correlation during training, with a well-chosen loss function. The result is a new latent space where we can directly modify the numerical value of attributes.

We show in qualitative and quantitative results that our method indeed gives disentangled editing results, while maintaining fidelity to other elements of the original image, such as face identity. Finally, we propose as simple and compact an architecture as possible for the autoencoder, in the interest of clarity and efficiency of the method. This is evidenced by our very short training times ($\sim$45 mins).



\section{Related work}

Given the great power of StyleGAN-type models, it is natural to manipulate its latent space $\mathcal{W}$ to achieve image editing. One of the first to propose such an approach was InterfaceGAN \cite{InterFaceGAN}, which finds linear editing directions by training a SVM in $\mathcal{W}$ for every single attribute. Following this idea, \cite{LatentTransformer} trains a network to determine a direction for every single latent vector in $\mathcal{W+}$.
Another approach, named GANSpace \cite{GANSpace}, proposes to apply PCA in the StyleGAN's latent space to analyse qualitatively its principal components and find attribute editing directions. This is done in an ad-hoc manner, and certain attributes may be missed.

A series of works~\cite{Image2StyleGAN,StyleFlow,Latent2Latent} have extended $\mathcal{W}$ to a space $\mathcal{W+}$ by applying a different style vector $w$ for each scale of the generator of StyleGAN. In effect, this allows editing to leave the original latent space $\mathcal{W}$ of StyleGAN. Image2StyleGAN~\cite{Image2StyleGAN} was the first to propose such an approach, which was then improved to Image2StyleGAN++~\cite{Image2StyleGAN++}. Other methods in $\mathcal{W+}$ have been proposed, such as StyleFlow~\cite{StyleFlow} which proposes to auto-encode latent vectors using a normalizing flow network or Latent2Latent~\cite{Latent2Latent} which modifies vectors using a dense network. Unfortunately, the use of the space $\mathcal{W+}$ is contrary to the goal of our work: indeed wish to stay in the same native latent space of the GAN. A further drawback of such an approach is that it only applies to style-type networks and cannot be generalised to any GAN. Also, many of these approaches use loss functions in both latent and image spaces. This implies high algorithmic costs during training (an image generation is required during training), which we wish to avoid. We also note that other spaces have been investigated such as the $\mathcal{S}$ space introduced by Wu \etal~\cite{StyleSpace} where the authors try to find which dimensions correspond to some given attributes with the help of segmentation masks.

Finally, we note that several StyleGAN encoders exist which project images into the $\mathcal{W+}$ latent space. Some of these include e4e~\cite{e4e} and pSp~\cite{pSp}.


\section{Method}

\noindent \textbf{StyleGAN2 and notation}: Since we use the StyleGAN2 model in our work, we first explain it here briefly, and also introduce some of our notation. To generate an image with StyleGAN2, a vector $z$ is sampled in a first latent space $\mathcal{Z}$ following a multidimensional normal distribution. This vector $z\in\mathcal{Z}$ is then transformed into a second space, denoted $\mathcal{W}$, passing through a mapping network $M:\mathcal{Z} \rightarrow \mathcal{W}$ made of an 8-layer MLP. The central idea of StyleGAN and StyleGAN2 is that this intermediate latent space $\mathcal{W}$ represents the ``style'' of images better than the original, probabilistic, latent space. The style is controlled by inserting the new vector $w=M(z)$, with $w \in \mathcal{W}$, into the generative network $G$, at different scales, in order to produce the final output image $y=G(w)$. The vector $w$ is inserted using an Adaptive Instance Normalization (AdaIn~\cite{huang2017arbitrary}). For a good visual illustration of the StyleGAN2 model, we refer the reader to the original paper~\cite{StyleGAN2}. An image synthesis using StyleGAN2 can thus be summarised as $y=G(M(z))$, with $z\sim \mathcal{N}(0,Id)$.

\begin{figure*}[t]
\begin{center}
\includegraphics[width=\textwidth]{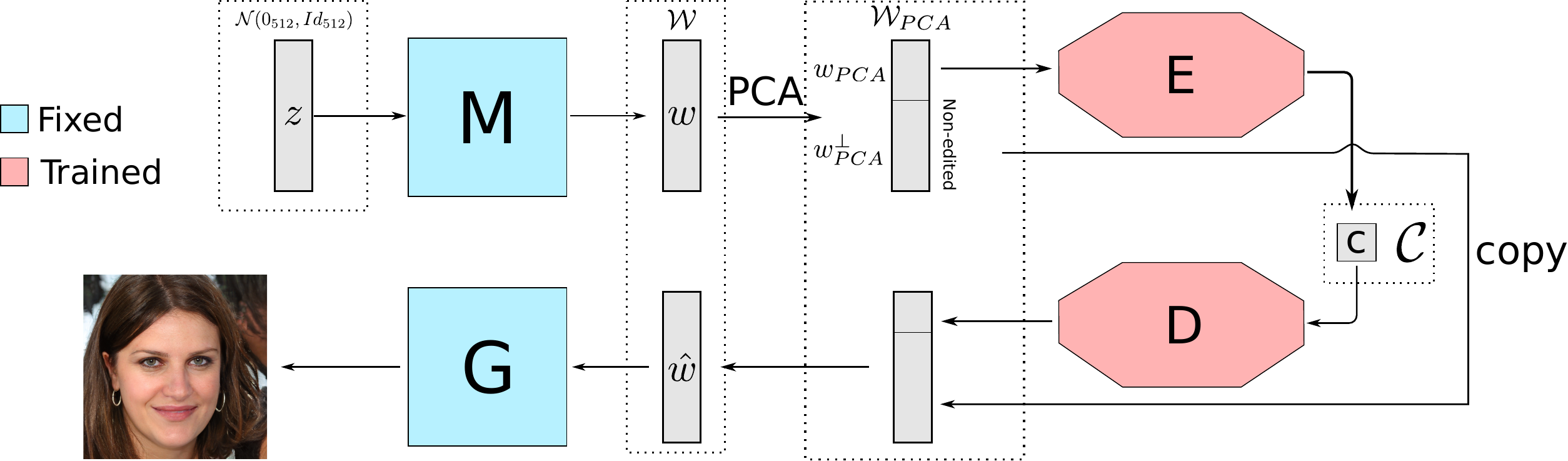}
\end{center}

\caption{Architecture of our editing pipeline for an image sampled from $\mathcal{Z} \sim \mathcal{N}(0_{512},I_{512})$. Note that the networks $M$ and $G$ are the pre-trained networks of StyleGAN2. The networks trained in our method are $E$ and $D$ (the autoencoder).}
\label{fig:archi}
\end{figure*}

\subsection{Creating an image/attribute database}
\label{subsec:imageAttributeDatabase}

The essence of our method is an autoencoder which transforms StyleGAN's latent space into another space where each attribute corresponds directly to one axis. In order to train such an autoencoder, we first require a labelled database of images to train it on. We generate a database of synthetic images using StyleGAN. Let $y=G(M(z))$ represent one such image. To label this image, we associate it with an attribute vector $a = (a_{1}, ...,a_{K})$. The attributes are determined with a pre-trained classification network $F$. In our case, we will take $K=40$ attributes. We then create a dataset $(w,a)$ by sampling in $\mathcal{Z}$, with:
\begin{equation} w = M(z) \mbox{ and } a = F \circ G(w).
\end{equation}

We can now proceed to describe our architecture.

\subsection{Architecture}

As mentioned above, the goal of our approach is to transform the latent space $\mathcal{W}$ such that each (labelled) semantic attribute of the generated image corresponds to a dimension of the transformed latent space. If we can achieve this, then editing becomes a simple matter of moving each axis independently, as a digital artist would move a slider. This disentanglement is crucial in image editing: the artist usually wants to be able to modify a single semantic factor in the image. 

\begin{figure}[t]
\begin{center}
\includegraphics[width=0.6\textwidth]{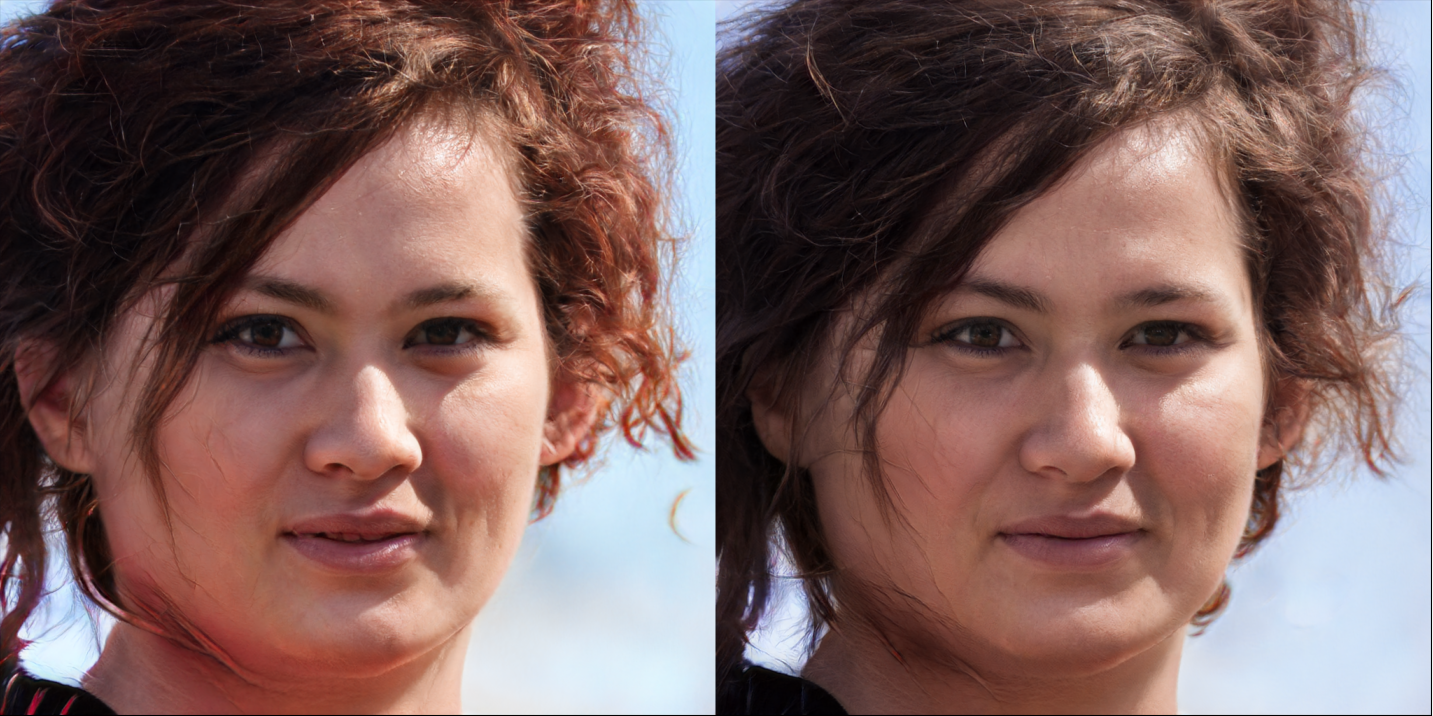}
\end{center}

\caption{Left: image generated with a $w \in \mathcal{W}$. Right: image generated with $w$ projected onto $\mathcal{W}_{PCA}$, corresponding to the 60 dimensions of greatest variability according to the PCA. We observe that the restriction to this sub-space has a very small influence on the resulting synthesised image.}

\label{fig:acp_shrunk}
\end{figure}


\subsubsection{A sub-space for extracting attributes}

Previous work~\cite{GANSpace} has noted that in $\mathcal{W}$, dimensions have an unequal influence on the generated images. Indeed, most GANs are deliberately over-parametrised. By applying the elbow method on the Principal Component Analysis (PCA) of $\mathcal{W}$, we can see that 60 dimensions are sufficient to correctly reconstruct an image and keep almost 80 \% of the space variability. We show an experiment to demonstrate this in Figure~\ref{fig:acp_shrunk}, where we have set the rest of the dimensions to 0. The impact of this operation is minimal, meaning that we can safely ignore these dimensions.

Thus, we decide to apply our auto-encoder in the sub-space corresponding to the first 60 dimensions of greatest variability, according to the PCA. We note this sub-space 
$\mathcal{W}_{PCA}$. This choice has two main advantages: it allows our auto-encoder to avoid performing editing trajectories in $\mathcal{W}$ dimensions of low variance, avoiding the appearance of artifacts and it also greatly reduces the parameter complexity of our auto-encoder. Let us note such a latent code transformed with the PCA as $\left[ w_{PCA}, w_{PCA}^{\perp}\right]$. We will not modify the second vector ${w_{PCA}}^{\perp}$, since we consider that it contains elements such as small texture variations. Note that we do \emph{not} set the elements of ${w_{PCA}}^{\perp}$ to 0 as in the experiment of Figure~\ref{fig:acp_shrunk}.

\subsubsection{An autoencoder for controllable attributes}

Contrary to the work of InterfaceGAN~\cite{InterFaceGAN} or GANspace~\cite{GANSpace}, which suppose that a linear direction is sufficient to achieve latent space attribute editing, we assume here that the editing path of an image can be more general. Thus, we propose to train an auto-encoder in $\mathcal{W}_{PCA}$ that extracts attributes from a latent code $w_{PCA}$, so that they can be modified individually in the manner of a slider. However, it is clear that $w_{PCA}$ contains more information than just the attributes; it will contain such elements as identity etc. Therefore, our autoencoder will project of code $w_{PCA}$ to a space where the first $K$ elements directly correspond to the desired attributes, and the rest are left free. In our case, we have $K = 40$ and $dim(\mathcal{C})=60$. Please note that the size of the latent space of our autoencoder is the same as $\mathcal{W}_{PCA}$: we do not compress latent codes, rather we re-organise them.

We will call the latent space of this auto-encoder $\mathcal{C}$ and $c = (c_{1}, \dots, c_{dim(\mathcal{C})})$ a latent vector in this space. We denote the encoder with $E$ and the decoder with $D$. Thus, $c=E(w_{PCA})$. Note that this definition assumes that $dim(\mathcal{C})>K$. This makes sense: an image must have greater latent dimensionality than its controllable attributes. The specific architecture of our autoencoder is given in Section~\ref{subsec:implementationDetails}.

In order to control the values of each attribute $a_k$, we set as a goal for our encoder $E$ to match the latent codes to the attributes: 

\begin{equation}
    c_{k} = a_{k}  \mbox{ for } k\in [\![ 1;K ]\!]
\end{equation}

The rest of the elements $c_{K+1,\dots,dim(\mathcal{C})}$ will not be modified, since we assume that it contains all other information which we do not want to modify (face identity, lighting etc).

To summarize, we carry out the following auto-encoding process: first we use PCA to project our vector $w \in \mathcal{W}$ into $\mathcal{W}_{PCA}$. In this way, we keep only the information strictly necessary for our network. Then, we proceed to the encoding of the vector $w_{PCA}$ by a multilayer perceptron to obtain a vector $c$. This vector represents a different attribute on the first $K$ dimensions and leaves the remaining dimensions free. At this point, we can edit the image attributes individually by modifying these first $K$ dimensions of $c$. Once this is done, the resulting vector is passed to the decoder $D$ to obtain $\hat w_{PCA}$. Finally, we apply the reverse operation of the PCA to return to the original space $\mathcal{W}$ and obtain our edited latent vector $\hat w$. A summary scheme of the architecture is proposed in Figure~\ref{fig:archi}.

\subsection{Training}

To train our auto-encoder, we use a latent code/attribute database. As mentioned in Section~\ref{subsec:imageAttributeDatabase}, this database can be built from a pre-trained classification network $F$.
For the loss function, we use a weighted sum of three terms during training:

\begin{itemize}

    \item A reconstruction term. Since we have an auto-encodeur, we want to be able to retrieve the original vector passed to our network if no modification is applied in the latent space $\mathcal{C}$. So we take the classical $\ell^2$ norm between input and outputs vectors for this task:
    \begin{equation} \mathcal{L}_{recons}(w_{PCA},\hat w_{PCA}) = ||w_{PCA} - \hat w_{PCA}||_{2}^{2} \end{equation}
    \item A loss function on the attributes. Since our goal is to edit image attributes, we want to fix the $c_{k}$ as being equal to $a_{k}$:
    \begin{equation} \mathcal{L}_{attr} =  \sum_{k=1}^{K} (a_{k}-c_{k})^{2} \end{equation}
    \item A correlation term. The motivation behind this term is to enforce the disentanglement of $\mathcal{C}$, which corresponds to independence between the dimensions of our latent space. For that, we try to set the autocorrelation matrix of the latent codes $Corr$ to a reference matrix $\Gamma_{ref}=I_{K}$:
    \begin{equation} \mathcal{L}_{corr} = \sum_{i,j}^{K,K} |Corr(i,j)-\Gamma_{ref}(i,j)| \end{equation} 
\end{itemize}
Note that the matrix $Corr$ is computed over a batch. Therefore, we need the batch size to be large enough during training in order to have a good estimate of this matrix. Additionally, since we aim at improving disentanglement, we set $\Gamma_{ref}=I_{K}$, but it
 is important to keep in mind that we can use another matrix if we would like the latent space to take the correlations into account.

Therefore, our final loss function is the following:
\begin{equation} \mathcal{L}_{total} = \mathcal{L}_{recons} + \alpha  \mathcal{L}_{attr} + \beta \mathcal{L}_{corr} \end{equation}

Please note that none of our loss functions require a forward pass of the generator to create an image, contrary to some of the previous works~\cite{Latent2Latent,StyleFlow}, whereas we carry out the entire training in the compressed space $\mathcal{W}_{PCA}$, greatly reducing training time (around 45 mins, see Table~\ref{tab:computationalImpact}).

 \section{Experiments and metrics}

 \subsection{Implementation details}
 \label{subsec:implementationDetails}

To train our network, we set $\alpha = 10^{-5}$ and $\beta = 10^{-5}$. Both encoder and decoder have the same architecture, an 8-layer MLP where each hidden layer is of size of 512, with the input and output size being of 60. Note that this indeed temporarily increases the size of the code, which we found gave good results. We use LeakyReLU activations, with parameter $10^{-2}$. Training is done for 150 epochs with a batch size of 256.

Our database is made of 200 000 latent vectors and their corresponding attributes which are obtained using an EfficientNet-B0\cite{EfficientNet} network pretrained on CelebA\cite{CelebA}.
 In general, the range of values for image attributes is $[0,1]$, which is made possible by the use of a sigmoid as the activation of the classifier network. Unfortunately, this leads to an under-representation of values close to $0.5$. Because of this, the network may learn a bimodal distribution for every latent code and might not be able to produce realistic results for $c_{k}$ near to $0.5$. To avoid such a behavior, we decide to "gaussianize" the attributes values of our database. In other words, for each attribute, we process a histogram equalization of its values and then apply the inverse cumulative distribution function of the normal law. In this way, we can always manipulate the attribute values in a known range and solve the issue.

\subsection{Our autoencoder variants}

As mentioned in the introduction, disentanglement of attributes is usually a key goal of image editing. However, in some situations, a digital artist may instead wish to impose certain correlations, which may come from observation of a database. This is completely possible in our method, since we directly control the attribute correlations via the loss function $\mathcal{L}_{corr}$. Indeed, we can set the reference correlation matrix $\Gamma_{ref}$ to whatever we choose.
Therefore, we present three variants of our approach:
\begin{itemize}
    \item variant A: no correlation loss is imposed. We use this as an ablation study of $\mathcal{L}_{corr}$;
    \item variant B: $\Gamma_{ref}$ is the correlation matrix of the attributes of the database;
    \item variant C: $\Gamma_{ref} = Id$ (disentanglement).
\end{itemize}
Variant B means that we accept the correlations present in the database.

\subsection{Qualitative evaluation}

\begin{figure}[h]
\begin{center}
\includegraphics[width=\textwidth]{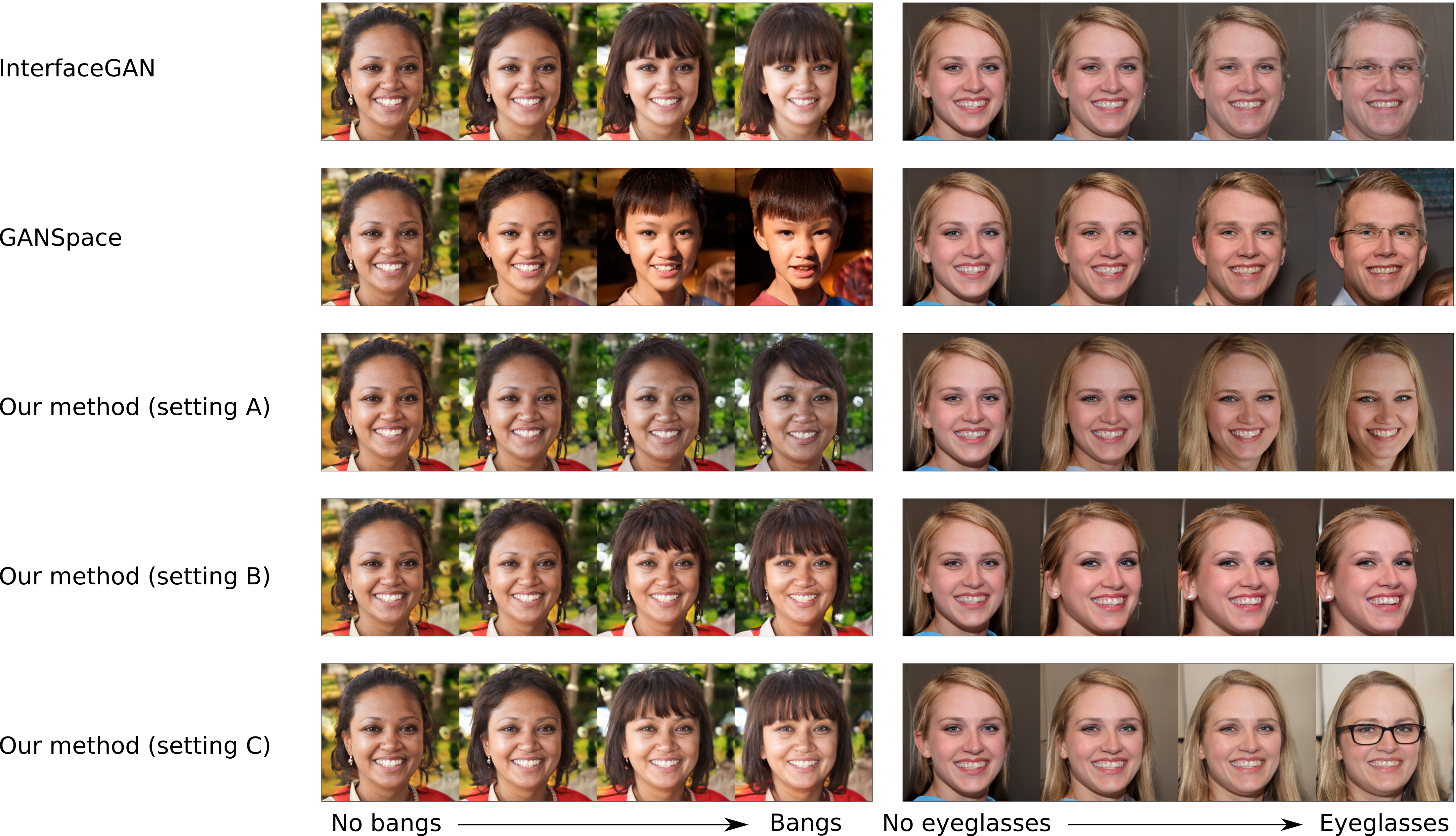}
\end{center}
\caption{Editing comparison between the variants of our method, InterfaceGAN and GANSpace. It is clear that variant C of our method achieves attribute editing in a disentangled manner. Variant B also achieves editing, but allows for correlated attributes.}
\label{fig:edits}
\end{figure}

\begin{figure}[h]
\begin{center}
\includegraphics[width=\textwidth]{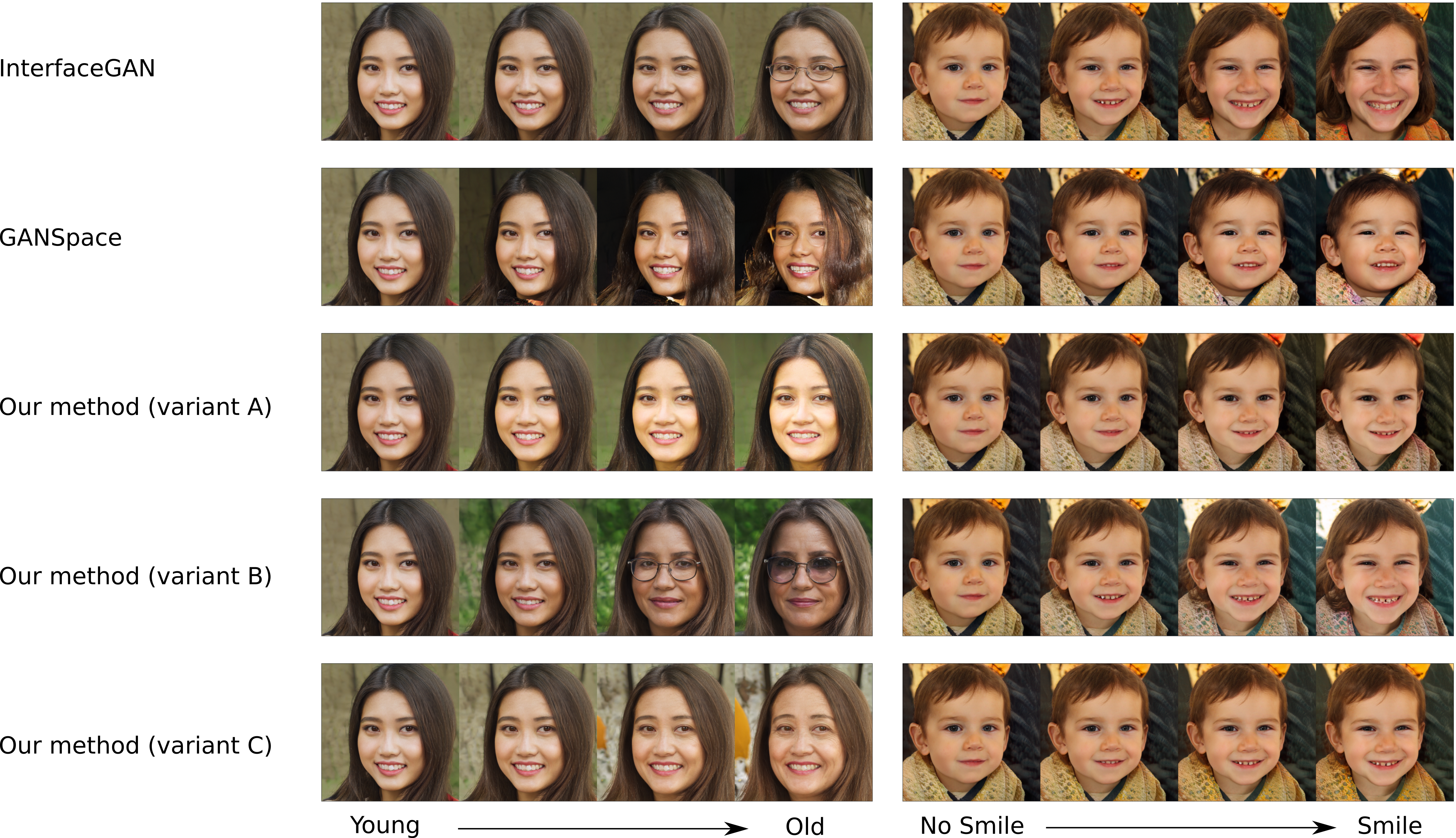}
\end{center}
\caption{Further editing comparison on the ``age'' and ``smile'' attributes. Again, it is clear that our variant C achieves the best disentanglement. Both InterfaceGAN and GANSpace heavily correlate attributes and modify the identity greatly.}
\label{fig:edits2}

\end{figure}

We compare our method to two previous works, InterfaceGAN~\cite{InterFaceGAN}  and GANSpace~\cite{GANSpace}, since they are the only state-of-the-art methods processing in $\mathcal{W}$, the latent space in which we work.
In the Figures \ref{fig:edits} \& \ref{fig:edits2}, we show visual comparisons between the three different variants of our method and the competing methods. It is clear that our editing gives much better disentanglement of attributes in the case of variant C, which is the desired behaviour.

\subsection{Quantitative evaluation}

Since our goal is to improve disentanglement while keeping an editing quality comparable to state-of-the-art methods, we use 3 different quantitative metrics to evaluate our method: identity conservation during an edit, attribute variation during an edit to quantify disentanglement and finally well-edited image rate (which is defined below). In this section, we only present the results of our method, variant C, since it corresponds to our initial goals.

To compute these three metrics for a given edited attribute $k$, we sample $N = 1024$ images in $\mathcal{W}$ and keep only samples with this attribute $k$ being negative: $(w_{1}^{k-}, ..., w_{N}^{k-})$. The attribute $k$ is then edited toward positive values. For this, we apply the edit at different amplitudes and we measure its target attribute value. When its value reaches 0.9 or above, we consider the edit done and keep the corresponding latent vectors: $(w_{1}^{k+}, ..., w_{N}^{k+})$. Thanks to these pairs of vectors, we can now measure the attributes and identity differences during the edit of attribute $k$.

In the interest of readability, we choose to apply our metrics to 7 out of 40 target attributes: {\it bangs}, {\it eyeglasses}, {\it gender}, {\it beard}, {\it skin tone}, {\it smile} and {\it age} (these attributes are common in the litterature). Note that for GANSpace, there is no principal component allowing us to edit skin tone \cite{GANSpace}. Hence, we did not take into consideration this attribute for our experiments on this method. 

\begin{table}[h]
\begin{center}
\begin{tabular}{ |p{3cm}||p{2cm}|p{2cm}|p{2cm}|  }
 \hline
 \multicolumn{4}{|c|}{Well-edited image rate ($\uparrow$)} \\
 \hline
Edited attribute & our method & InterfaceGAN & GANSpace \\
 \hline
 Bangs        & 0.924 & {\bf0.980} & 0.635 \\
 Eyeglasses   &  {\bf0.999} &  0.995 & 0.933 \\
 Gender       &   0.975  &  0.979 & {\bf0.990} \\
 No\_Beard    & 0.997 & {\bf1.} & {\bf 1.} \\
 Pale\_Skin   & {\bf0.902} & 0.887 & XXX \\
 Smile        & {\bf 1.} & 0.998 & 0.906 \\
 Age          &  0.950 & {\bf0.961} & 0.603 \\
 \hline
\end{tabular}
\end{center}
\label{tab:well-edited}
\caption{Percentage of well-edited images for our method and other approaches.}
\end{table}

\begin{figure}[h]
\begin{center}
\includegraphics[width=\textwidth]{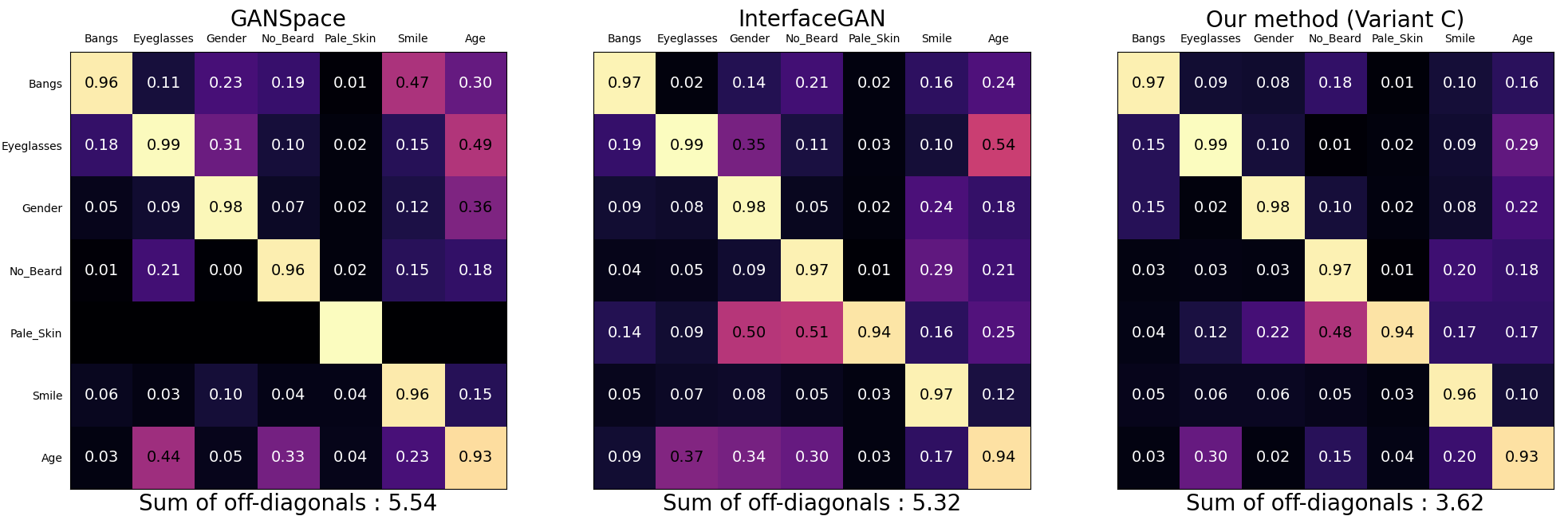}
\end{center}
\caption{Attribute variation matrices. Each row corresponds to the edit of an attribute and indicates the variations of all the attributes. In other words, the coefficient $(k,l)$ corresponds the variation of the attribute $l$ when we edit the $k^{th}$ one. We see that our method outperforms the others in terms of off-diagonal sum (which can be seen as a meaure of entanglement). Note that for all sums of off-diagonal coefficients, we did not take into account the Pale\_Skin row which GANSpace is not able to modify.}
\label{fig:attr_matrix}
\end{figure}

Since disentanglement requires that a single semantic factor is modified during an edit, our goal is to minimize all attributes variations except the target one.
To visualize this property, we plot a matrix (figure \ref{fig:attr_matrix}) where each coefficient corresponds to the average variation of the column-attribute along a row-attribute edit:

\begin{equation} Mat_{k,l} = {{1}\over{N}} \sum_{i=1}^{N} (F_{l} \circ G(w_{i}^{k+}) - F_{l} \circ G(w_{i}^{k-})) \end{equation}
where $F_{l}$ is the the network $F$'s estimate of the attribute $l$ value.

For the well-edited image rate, we simply take the rate of images that have reached a 0.9 target attribute value along the edits. This quantity is shown at table \ref{tab:well-edited} where we can see that our proposed method is as good as InterfaceGAN and GANSpace.

Furthermore we show in appendix that our network allows us to preserve the identity equivalently to state-of-the-art methods. Hence, the proposed method improves the disentanglement property while keeping an equivalent fidelity of editions. 
Finally, we provide in table \ref{tab:computationalImpact} some informations about the computational ressources needed to train our network.

\vspace{5pt}

\begin{table}[h]
\begin{center}
\begin{tabular}{ |p{4cm}|p{2cm}|p{3cm}|  }
 \hline
 \multicolumn{3}{|c|}{Computational impact (on a Nvidia RTX3090)} \\
 \hline
Number of parameters & Training time & Energy consumption \\
 \hline
 3~553~520 & 45 min 36s & 0.128 kW.h \\
 \hline
\end{tabular}
\end{center}
\caption{Computational impact and complexity of our model}
\label{tab:computationalImpact}
\end{table}

\section{Conclusion}

We have proposed a method to edit facial image attributes which uses an autoencoder applied to the latent space of StyleGAN2. This autoencoder transforms the latent space into another space in which attributes correspond to a single axis, allowing for direct editing, in the manner of a slider for each attribute. Furthermore we encourage these axes to be decorrelated, ensuring a disentangled representation. These goals are achieved by two well- chosen loss functions. Furthermore, we encourage image fidelity by working in a sub-space of the latent space, found using a PCA. We have shown qualitative and quantitative results which demonstrate that our method indeed edits the desired attributes in a disentangled fashion, while maintainting fidelity to the identity of the person. Our autoencoder is quite compact, with around 3.5 million parameters, and is very straightforward to implement.

\section*{Acknowledgements}
This work was granted access to the HPC resources of IDRIS under the allocation 20XX-AD011013749R1 made by GENCI.

We acknowledge the support of the French Agence Nationale de la Recherche (ANR) under reference ANR-21-CE23-0024 IDEGEN.

\bibliographystyle{unsrt}  
\bibliography{main}  

\begin{thebibliography}{10}

\bibitem{kingma2013auto}
Diederik~P Kingma and Max Welling.
\newblock Auto-encoding variational bayes.
\newblock {\em arXiv preprint arXiv:1312.6114}, 2013.

\bibitem{papamakarios2021normalizing}
George Papamakarios, Eric Nalisnick, Danilo~Jimenez Rezende, Shakir Mohamed,
  and Balaji Lakshminarayanan.
\newblock Normalizing flows for probabilistic modeling and inference.
\newblock {\em The Journal of Machine Learning Research}, 22(1):2617--2680,
  2021.

\bibitem{song2020score}
Yang Song, Jascha Sohl-Dickstein, Diederik~P Kingma, Abhishek Kumar, Stefano
  Ermon, and Ben Poole.
\newblock Score-based generative modeling through stochastic differential
  equations.
\newblock {\em arXiv preprint arXiv:2011.13456}, 2020.

\bibitem{goodfellow2020generative}
Ian Goodfellow, Jean Pouget-Abadie, Mehdi Mirza, Bing Xu, David Warde-Farley,
  Sherjil Ozair, Aaron Courville, and Yoshua Bengio.
\newblock Generative adversarial networks.
\newblock {\em Communications of the ACM}, 63(11):139--144, 2020.

\bibitem{BigGAN}
Andrew Brock, Jeff Donahue, and Karen Simonyan.
\newblock Large scale {GAN} training for high fidelity natural image synthesis.
\newblock {\em CoRR}, abs/1809.11096, 2018.

\bibitem{StyleGAN}
Tero Karras, Samuli Laine, and Timo Aila.
\newblock A style-based generator architecture for generative adversarial
  networks.
\newblock {\em CoRR}, abs/1812.04948, 2018.

\bibitem{StyleGAN2}
Tero Karras, Samuli Laine, Miika Aittala, Janne Hellsten, Jaakko Lehtinen, and
  Timo Aila.
\newblock Analyzing and improving the image quality of stylegan.
\newblock {\em CoRR}, abs/1912.04958, 2019.

\bibitem{GANSpace}
Erik H{\"{a}}rk{\"{o}}nen, Aaron Hertzmann, Jaakko Lehtinen, and Sylvain Paris.
\newblock Ganspace: Discovering interpretable {GAN} controls.
\newblock {\em CoRR}, abs/2004.02546, 2020.

\bibitem{InterFaceGAN}
Yujun Shen, Ceyuan Yang, Xiaoou Tang, and Bolei Zhou.
\newblock Interfacegan: Interpreting the disentangled face representation
  learned by gans.
\newblock {\em CoRR}, abs/2005.09635, 2020.

\bibitem{LatentTransformer}
Xu~Yao, Alasdair Newson, Yann Gousseau, and Pierre Hellier.
\newblock A latent transformer for disentangled and identity-preserving face
  editing.
\newblock {\em CoRR}, abs/2106.11895, 2021.

\bibitem{Image2StyleGAN}
Rameen Abdal, Yipeng Qin, and Peter Wonka.
\newblock Image2stylegan: How to embed images into the stylegan latent space?
\newblock {\em CoRR}, abs/1904.03189, 2019.

\bibitem{StyleFlow}
Rameen Abdal, Peihao Zhu, Niloy~J. Mitra, and Peter Wonka.
\newblock Styleflow: Attribute-conditioned exploration of stylegan-generated
  images using conditional continuous normalizing flows.
\newblock {\em CoRR}, abs/2008.02401, 2020.

\bibitem{Latent2Latent}
Siavash Khodadadeh, Shabnam Ghadar, Saeid Motiian, Wei-An Lin, Ladislau
  Bölöni, and Ratheesh Kalarot.
\newblock Latent to latent: A learned mapper for identity preserving editing of
  multiple face attributes in stylegan-generated images.
\newblock In {\em 2022 IEEE/CVF Winter Conference on Applications of Computer
  Vision (WACV)}, pages 3677--3685, 2022.

\bibitem{Image2StyleGAN++}
Rameen Abdal, Yipeng Qin, and Peter Wonka.
\newblock Image2stylegan++: How to edit the embedded images?
\newblock {\em CoRR}, abs/1911.11544, 2019.

\bibitem{StyleSpace}
Zongze Wu, Dani Lischinski, and Eli Shechtman.
\newblock Stylespace analysis: Disentangled controls for stylegan image
  generation.
\newblock {\em CoRR}, abs/2011.12799, 2020.

\bibitem{e4e}
Omer Tov, Yuval Alaluf, Yotam Nitzan, Or~Patashnik, and Daniel Cohen{-}Or.
\newblock Designing an encoder for stylegan image manipulation.
\newblock {\em CoRR}, abs/2102.02766, 2021.

\bibitem{pSp}
Elad Richardson, Yuval Alaluf, Or~Patashnik, Yotam Nitzan, Yaniv Azar, Stav
  Shapiro, and Daniel Cohen{-}Or.
\newblock Encoding in style: a stylegan encoder for image-to-image translation.
\newblock {\em CoRR}, abs/2008.00951, 2020.

\bibitem{huang2017arbitrary}
Xun Huang and Serge Belongie.
\newblock Arbitrary style transfer in real-time with adaptive instance
  normalization.
\newblock In {\em Proceedings of the IEEE international conference on computer
  vision}, pages 1501--1510, 2017.

\bibitem{EfficientNet}
Mingxing Tan and Quoc~V. Le.
\newblock Efficientnet: Rethinking model scaling for convolutional neural
  networks.
\newblock {\em CoRR}, abs/1905.11946, 2019.

\bibitem{CelebA}
Ziwei Liu, Ping Luo, Xiaogang Wang, and Xiaoou Tang.
\newblock Deep learning face attributes in the wild.
\newblock In {\em Proceedings of International Conference on Computer Vision
  (ICCV)}, December 2015.

\bibitem{face-recognition}
Adam Geitgey.
\newblock face\_recognition python package.
\newblock \url{https://github.com/ageitgey/face_recognition}, 2021.

\bibitem{FID}
Martin Heusel, Hubert Ramsauer, Thomas Unterthiner, Bernhard Nessler,
  G{\"{u}}nter Klambauer, and Sepp Hochreiter.
\newblock Gans trained by a two time-scale update rule converge to a nash
  equilibrium.
\newblock {\em CoRR}, abs/1706.08500, 2017.

\end{thebibliography}

\newpage

\section{Appendix}
\subsection{Identity preservation}

To quantify identity preservation for a given attribute edit, we use the library \cite{face-recognition} which consist in embbeding a face image into the hidden layers of a face recognition network $H$. Hence, the identity similarity is set as the cosine distane between the embeddings:
\begin{equation} D_{id}^{k} = {{1}\over{N}} \sum_{i=1}^{N} {{<H(w_{i}^{k-})|H(w_{i}^{k+})>}\over{||H(w_{i}^{k-})||.||H(w_{i}^{k+})||}} \end{equation}

We recall that there is no satisfying direction for editing the 'Pale\_skin' attribute with GANSpace in $\mathcal{W}$.

\begin{table}[h]
\begin{center}
\begin{tabular}{ |p{3cm}||p{2cm}|p{2cm}|p{2cm}|  }
 \hline
 \multicolumn{4}{|c|}{Identity preservation ($\uparrow$)} \\
 \hline
Edited attribute & InterfaceGAN & GANSpace & our method \\
 \hline
 Bangs        & 0.949 & 0.907 & {\bf 0.950}  \\
 Eyeglasses   & 0.932 & 0.917 & {\bf 0.950} \\
 Gender       & {\bf 0.939} & 0.933 & 0.926 \\
 No\_Beard    & {\bf 0.951} & 0.950 &  0.945 \\
 Pale\_Skin   &  0.895 & XXX & {\bf 0.908} \\
 Smile        & 0.964 & 0.930 & {\bf 0.968} \\
 Age          & 0.900 & {\bf 0.910} & {\bf 0.910} \\
 \hline
\end{tabular}
\end{center}
\label{tab:identity}
\caption{Identity preservation between original images and attribute-edited ones for each method.}
\end{table}

As we can see in the table \ref{tab:identity}, our method achieves equivalent results compared to InterfaceGAN while outperforming GANSpace.

\subsection{FID}

In order to confirm that our method achieves at least equal edit stability compared to InterfaceGAN and GANSpace, we also propose here to compute the Frechet Inception Distance \cite{FID} (also known as FID) between the two sets of images (original and edited ones):

\begin{equation} D_{FID}^{k} = FID( \{ w_{i}^{k-}| \forall i \in [\![ 1,N ]\!] \} , \{w_{i}^{k+}| \forall i \in [\![ 1,N ]\!] \} ) \end{equation}

Indeed, since FID consist in quantifying the average visual difference between two sets of images, it means that the better FID, the closer to original images we are. In other words, the editing method introduces less perceptual changes while achieving the desired edit.

\begin{table}[h]
\begin{center}
\begin{tabular}{ |p{3cm}||p{2cm}|p{2cm}|p{2cm}|  }
 \hline
 \multicolumn{4}{|c|}{FID ($\downarrow$) } \\
 \hline
Edited attribute & InterfaceGAN & GANSpace & our method \\
 \hline
 Bangs        & 67.6 & 137.9 & {\bf 62.9} \\
 Eyeglasses   & 53.0 & 85.0 & {\bf 44.9} \\
 Gender       & 67.5 & 108.1 & {\bf 65.4} \\
 No\_Beard    & 50.5 & {\bf 48.1} & 52.2 \\
 Pale\_Skin   & 98.9 & XXX & {\bf 86.8} \\
 Smile        & 29.8 & 67.0 & {\bf 26.6} \\
 Age          & 67.6. & 85.1 & {\bf 42.4} \\
 \hline
\end{tabular}
\end{center}
\label{tab:fid}
\caption{FID between original images and attribute-edited ones for each method.}
\end{table}


\end{document}